# Comparative Study of Deep Learning Architectures for Textual Damage Level Classification

Aziida Nanyonga
School of Engineering and Technology
University of New South Wales
Canberra, Australia
a.nanyonga@unsw.edu.au

Hassan Wasswa
School of Systems and Computing
University of New South Wales
Canberra, Australia
h.wasswa@unsw.edu.au

Graham Wild
School of Science
University of New South Wales
Canberra, Australia
g.wild@unsw.edu.au

*Abstract*— Given the paramount importance of safety in the aviation industry, even minor operational anomalies can have significant consequences. Comprehensive documentation of incidents and accidents serves to identify root causes and propose safety measures. However, the unstructured nature of incident event narratives poses a challenge for computer systems to interpret. Our study aimed to leverage Natural Language Processing (NLP) and deep learning models to analyze these narratives and classify the aircraft damage level incurred during safety occurrences. Through the implementation of LSTM, BLSTM, GRU, and sRNN deep learning models, our research yielded promising results, with all models showcasing competitive performance, achieving an accuracy of over 88% significantly surpassing the 25% random guess threshold for a four-class classification problem. Notably, the sRNN model emerged as the top performer in terms of recall and accuracy, boasting a remarkable 89%. These findings underscore the potential of NLP and deep learning models in extracting actionable insights from unstructured text narratives, particularly in evaluating the extent of aircraft damage within the realm of aviation safety occurrences.

## I. INTRODUCTION

In aviation safety, timely and accurate classification of aircraft damage levels is critical for effective incident analysis and the development of preventive measures [1]. Accurate damage level classification facilitates the assessment of safety risks associated with aviation incidents and accidents, aiding in the enhancement of proactive safety protocols and measures [2]. To address the challenges of accurately categorizing damage levels, we leverage the power of deep learning models in processing and analyzing textual narratives related to aviation incidents. Our study focuses on the classification of damage levels using the prominent ATSB (Air Transport Safety Bureau) dataset, a comprehensive collection of textual reports detailing aviation incidents and associated damage levels.

The accurate classification of damage levels in aviation incidents is essential for various stakeholders, including aviation regulatory authorities, airlines, manufacturers, and safety analysts. Understanding the severity of damage incurred during aviation incidents facilitates the development of robust safety protocols, maintenance practices, and incident response strategies [3], [4]. Additionally, the efficient categorization of damage levels enables the identification of recurrent patterns and potential areas of improvement in aircraft design, maintenance procedures, and operational protocols [5], [6]. Consequently, our study contributes to the advancement of aviation safety practices and plays a pivotal role in fostering a culture of proactive safety measures within the aviation industry.

The motivation behind this research stems from the necessity to improve the current methods of damage level classification in the aviation industry. While traditional approaches rely heavily on manual inspection and analysis, our study aims to streamline and automate the classification process using advanced deep-learning models. By leveraging natural language processing (NLP) techniques and robust deep learning architectures, we seek to enhance the accuracy and efficiency of damage-level classification, enabling a more precise understanding of the severity of aviation incidents. This, in turn, empowers aviation stakeholders to make data-driven decisions and implement targeted safety measures, leading to an overall reduction in aviation risks and enhanced passenger safety.

The primary objective of this study is to evaluate the efficacy of various deep learning models, including Long Short-Term Memory (LSTM), Simple Recurrent Neural Network (sRNN), Bidirectional LSTM (BLSTM), and Gated Recurrent Unit (GRU), in accurately classifying textual narratives associated with different levels of aircraft damage. Through a comparative analysis of these models, we aim to identify the most effective and efficient approach for classifying damage levels, considering factors such as precision, recall, F1-score, and accuracy. Additionally, our study seeks to provide valuable insights into the performance and applicability of different deep-learning architectures in the domain of aviation safety, thereby contributing to the advancement of automated incident analysis and proactive safety management practices in the aviation industry.

The structure of this paper is as follows: Section II provides a review of the existing literature, highlighting the significance of aircraft damage level classification in aviation safety research and discussing relevant prior work. Section III gives an account of the methodology employed in this study, including data preprocessing, model selection, training, and evaluation. Section IV presents the results of our experiments, showcasing the performance of the deep learning models in damage level classification and discussing the interpretation of results, potential limitations, and implications for aviation safety. Finally, Section V concludes the paper by summarizing the key findings with suggestions for potential future research directions in the field of aviation safety management and incident analysis.

## II. RELATED WORK

The study of damage level classification in the context of aviation incidents has garnered considerable attention in recent years, with a growing emphasis on leveraging NLP techniques and machine learning approaches for effective analysis. Notably, various studies have focused on the application of machine learning algorithms for incident severity classification, enabling the identification of critical patterns and trends in aviation safety data [7], [8].

In the specific domain of NLP-based incident analysis, researchers have explored the utilization of different text-processing methodologies, including sentiment analysis, topic modelling, and classification techniques, to extract valuable insights from textual data [9], [10]. Moreover, the application of deep learning models, such as LSTM, GRU, and other recurrent neural networks, has gained prominence in various NLP tasks, showcasing promising results in text classification and sentiment analysis [7], [11]. Another study [12] conducted an extensive exploration of various RNN architectures for sentence modelling, affirming the suitability of these models for sequential datasets like text mining. Additionally, Pang et al. [13] applied RNN techniques to predict weather-related tasks, effectively regulating pre-flight information. Paul's comprehensive review of NLP tools in civil aviation emphasized the potential of RNN, suggesting its efficacy in mining time series data [8]. Chanen [14] proposed a deep learning approach to extract meaningful narratives from aviation safety reports, utilizing a word2vec model for semantic analysis in 186,000 ASRS reports, thereby enhancing the interpretability of safety documentation for experts. Zhang et al. [15] focused on aviation safety prognosis, employing LSTM and word embeddings to classify NTSB reports, highlighting the utility of deep learning models in improving safety analysis. Furthermore, ElSaid et al. [16] addressed the prediction of excess events in aircraft engines using LSTM recurrent neural networks, demonstrating the superior predictive capabilities of LSTM over traditional RNN architectures, particularly in the context of flight vibration datasets.

Research in aviation safety has employed various techniques to understand safety occurrences. Nanyonga et al. focused on the classification of aviation safety occurrences using natural language processing (NLP) and AI models. The study aimed to infer the damage level to the aircraft from text narratives. Evaluating the performance of various deep learning models including LSTM, BLSTM, GRU, and sRNN, they analyzed a dataset of 27,000 safety occurrence reports from the NTSB. Their results indicated competitive performance across all models. Meanwhile, Inan, [2] study delved into aircraft damage classification using machine learning methods, highlighting the impact of various factors such as zones, weather, time, and historical context on civil aviation incidents. They employed a set of machine learning algorithms, including logistic regression (LR), artificial neural networks (ANN), and decision trees (DT), to assess the significance of different parameters in classifying aircraft damage.

Additionally, Nick et al. [6] utilized an agent-based structural health monitoring system, employing unsupervised learning for identifying the existence and location of damage and supervised learning for identifying the type and severity of damage. The supervised learning techniques included support vector machines (SVM), naive Bayes classifiers (NB), and feed-forward neural networks (FFNN), while unsupervised learning techniques encompass k-means and self-organizing maps (SOM).

Furthermore, recent studies have highlighted the potential of deep learning architectures in enhancing the accuracy and efficiency of incident severity classification, demonstrating their capability to handle complex textual data and capture nuanced patterns in incident narratives [17], [18]. However, a comparative analysis of multiple deep learning models for damage level classification in the aviation domain, especially using the ATSB dataset, remains relatively scarce in the current literature.

Despite the existing research efforts, a critical gap persists in the identification of the most effective deep learning approach for damage level classification, particularly in the context of aviation incident narratives. This study aims to bridge this gap by conducting a detailed comparative analysis of LSTM, SRNN, BLSTM, and GRU models, providing valuable insights into the performance and applicability of these models in aviation incident analysis.

### III. METHODOLOGY

In this section, we outline the methodology employed in this research to classify aircraft damage levels within safety occurrence reports from the ATSB using NLP and Deep Learning techniques. Our approach involves data preprocessing, model selection, training, and evaluation as depicted in Fig. 1.

*A. Data Collection*

Various organizations such as the Aviation Safety Reporting System (ASRS), the National Transportation Safety Board (NTSB) and ATSB collect and publish aviation incident investigation reports. For this study, the researchers utilized the ATSB aviation incident reports that were recorded in Australia for 10 years, resulting in a dataset with 53,275 records where the data was sourced directly from the ATSB investigation authorities spanning from 1/01/2013 to 12/31/2022. Moreover, the dataset comprised 50,778 records following data preprocessing and cleaning. From each report, the 'Summary' and 'damageLevel' fields were extracted for training and validation of deep learning models.

*B. Text Processing*

Text preprocessing plays a crucial role in preparing unstructured text data for machine learning models. In this study, we leveraged the Keras deep learning library for its extensive collection of deep learning models and model layers. The Tokenizer module was utilized to efficiently generate tokens and sequence vectors from input text. To encode categorical data, such as DamageLevel labels (i.e., destroyed, Substantial, Minor, and None), we employed the to_categorical module in Keras, mapping these categorical entries to numerical values using one-hot encoding for each data instance.

To address challenges related to special characters, punctuation, and stop words, as well as to perform word lemmatization, we harnessed the capabilities of the spacy library. Spacy is a Python library tailored for text-processing tasks, encompassing functionalities like named entity recognition and word tagging. It maintains an extensive list of special characters, punctuation marks, and stop words and undergoes regular updates whenever necessary to remain current.

With the aforementioned tools at our disposal, each narrative underwent a comprehensive preprocessing pipeline, ultimately being transformed into a representative sequence or vector with a fixed length of 2000. For narratives with fewer than 2000 words, we padded the numeric sequences with zeros, ensuring uniformity. In contrast, narratives exceeding 2000 words were truncated to meet this standardized length. The vocabulary size of the text corpus was set to 100,000, accommodating a broad range of terms.

To partition the dataset into training (80%), and testing (20%) sets, we utilized the train-test-split module from scikit-learn. All experiments conducted in this study were implemented using the Python programming language, with Jupyter Notebook serving as the chosen code editor. This rigorous text preprocessing framework laid the foundation for subsequent model training and evaluation, enabling the accurate classification of aircraft damage levels based on unstructured safety occurrence narratives from the ATSB dataset.

*C. Text Classification*

To ensure model robustness and prevent overfitting during the training, 10% of the train-set was set aside for model validation in each training epoch. This practice facilitated continuous evaluation and refinement of the models. Deep learning models including sRNN, LSTM, BLSTM, and GRU are trained on this data, each offering unique capabilities for text classification tasks. Model optimization was accomplished using the Adam optimizer, chosen for its efficiency in gradient-based optimization [19]. It is noteworthy that this study did not focus explicitly on identifying the best optimizer, thus allowing for the exploration of alternative optimization techniques in future research endeavours.

*D. Deep Learning Model Architecture*

For consistency and comparability across all models, a shared architecture served as the foundation, with slight adjustments for each model. This standardized architecture comprised three key components: an embedding layer, hidden layers, and an output layer. To introduce non-linearity and capture complex relationships in the data, the Rectified Linear Unit (ReLU) activation function was applied to all hidden layers. Meanwhile, the SoftMax activation function was adopted for the output layer, facilitating multi-class classification. The final predicted class was determined using the argmax function, which identifies the index associated with the highest probability in the SoftMax output. For a visual representation of the deep learning architectures employed in this study, please refer to Fig. 2.

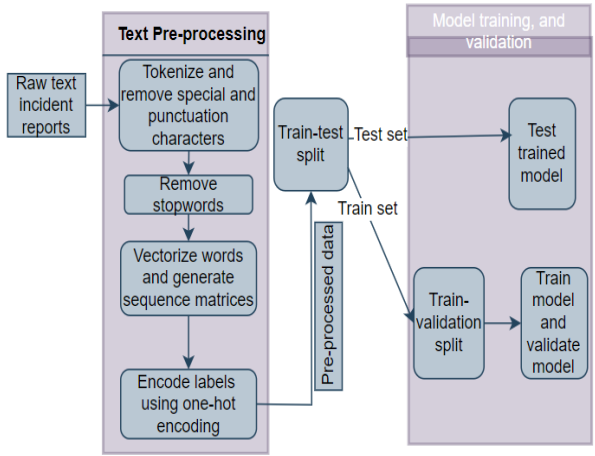

Fig. 1. Methodological framework

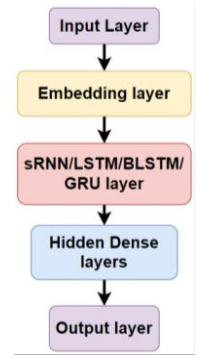

Fig. 2. Deep learning architectures

This consistent architecture provided a solid foundation for training and evaluation of the deep learning models, enabling a fair comparison of their performance and the accurate classification of aircraft damage levels based on unstructured safety occurrence narratives.

*1) Simple Recurrent Neural Networks (RNNs).*

The Simple Recurrent Neural Network (sRNN) is a fundamental type of RNN architecture that processes sequential data by feeding the output of the previous time step as input to the current time step. Its architecture is relatively basic, consisting of a single hidden layer that facilitates the flow of information from one step to the next [20]. The mathematical formulation for an sRNN can be described as follows:

$h_t = \sigma(W_h \cdot [h_{t-1}, x_t] + b_h)$, where $h_t$ represents the hidden state at time step $t$, $x_t$ denotes the input at time step $t$, and $\sigma$ is the activation function. Despite its simplicity, the sRNN may struggle to capture long-range dependencies and complex temporal patterns, making it more suitable for simple sequential tasks that do not involve intricate contextual relationships.

*2) GRU (Gated Recurrent Unit)*

The Gated Recurrent Unit (GRU) is a type of recurrent neural network (RNN) that operates on sequential data. It was designed to address the limitations of the vanishing gradient problem and the computational cost associated with the more complex LSTM architecture. GRU achieves this through a simplified architecture that consists of two gates: the update gate and the reset gate. These gates control the flow of information within the network and regulate the retention or discarding of information from the previous time step. The update gate determines how much of the past information needs to be carried forward, while the reset gate decides which parts of the past information should be forgotten [21]. The architecture of the GRU can be expressed mathematically through the following formulas:

- Update gate : $z_t = \sigma(W_z \cdot [h_{t-1}, x_t])$ , where $z_t$ represents the update gate output at time step $t$, $h_{t-1}$ is the previous hidden state, $x_t$ is the input at time step $t$, and $W_z$ is the weight matrix associated with the update gate.
- Reset gate: $r_t = \sigma(W_r \cdot [h_{t-1}, x_t])$ , where $r_t$ represents the reset gate output at time step $t$ , and $W_r$ is the weight matrix associated with the reset gate.
- New memory content: $h'_t = tanh(W \cdot [r_t \odot h_{t-1}, x_t])$ , where $h'_t$ represents the new candidate

hidden state at time step $t$, and $W$ is the weight matrix associated with the new memory content.
- Final memory content: $h_t = (1 - z_t) \odot h_{t-1} + z_t \odot h'_t$, where $h_t$ represents the updated hidden state at time step $t$, combining the previous hidden state with the new memory content based on the update gate output.

*3) LSTM (Long Short-Term Memory)*

The Long Short-Term Memory (LSTM) network is a type of RNN that was developed to address the issue of capturing long-range dependencies in sequential data. It is designed to store information over long periods, making it well-suited for tasks that involve understanding context and temporal patterns. The architecture of LSTM is more complex compared to that of a basic RNN, featuring memory cells and various gates, including the input gate, output gate, and forget gate [11]. These gates regulate the flow of information and determine which information to keep or discard. The key formulas involved in the LSTM architecture are as follows:

- Forget gate: $f_t = \sigma(W_f \cdot [h_{t-1}, x_t] + b_f)$, where $f_t$ represents the output of the forget gate at time step $t$, $h_{t-1}$ is the previous hidden state, $x_t$ is the input at time step $t$, and $W_f$ is the weight matrix associated with the forget gate, and $b_f$ is the bias.
- Input gate: $i_t = \sigma(W_i \cdot [h_{t-1}, x_t] + b_i)$, where $i_t$ represents the output of the input gate at time step $t$, $W_i$ is the weight matrix associated with the input gate, and $b_i$ is the bias.
- New candidate values: $g_t = tanh(W_g \cdot [h_{t-1}, x_t] + b_g)$, where $g_t$ represents the new candidate values at time step $t$, $W_g$ is the weight matrix associated with the candidate values, and $b_g$ is the bias.
- Cell state: $C_t = f_t \odot C_{t-1} + i_t \odot g_t$, where $C_t$ represents the cell state at time step $t$, combining the previous cell state with the input and forget gate outputs.
- Output gate: $o_t = \sigma(W_o \cdot [h_{t-1}, x_t] + b_o)$, where $o_t$ represents the output of the output gate at time step $t$, $W_o$ is the weight matrix associated with the output gate, and $b_o$ is the bias.
- Hidden state: $h_t = o_t \odot tanh(C_t)$, where $h_t$ represents the hidden state at time step $t$, combining the cell state with the output gate output.

*4) BLSTM (Bidirectional Long Short-Term Memory)*

The Bidirectional Long Short-Term Memory (BLSTM) model is an extension of the traditional LSTM architecture that processes input sequences in both forward and backward directions. By incorporating bidirectional processing, BLSTM can capture information from both past and future contexts simultaneously. This allows the model to understand the context and dependencies in a more comprehensive manner, making it well-suited for tasks that require a deep understanding of the sequence context. The mathematical formulation for a BLSTM includes the combination of forward and backward LSTM operations, allowing the model to capture long-term dependencies effectively [22]. BLSTM is commonly used in tasks such as speech recognition, language translation, and named entity recognition, where understanding both past and future context is critical for accurate predictions and analyses.

*E. Model Performance Evaluation*

This section elucidates the evaluation criteria utilized in this study to assess the models' performance. The primary focus of this research is multi-class classification, and as such, performance was gauged based on the accuracy of predictions across various classes. To comprehensively evaluate model performance, we employed a suite of standard prediction performance metrics, including recall, F1-score, precision, and accuracy.

## IV. RESULTS AND DISCUSSION

In this study, we have explored the application of NLP and Deep Learning techniques to enhance aviation safety analysis by classifying aircraft damage level within safety occurrence reports. Leveraging a substantial dataset of 50,778 safety occurrence reports provided by ATSB, we evaluated the performance of advanced Deep Learning models, including sRNN, LSTM, BLSTM, and, GRU using a variety of performance metrics.

*A. Model Performance*

Our findings reveal the remarkable capabilities of these models in accurately classifying damage level of aircraft within unstructured text narratives. Table III shows the key performance results for each model.

TABLE I. DEEP LEARNING MODEL PERFORMANCE

| Models | Precision (%) | Recall (%) | F1 (%) | Accuracy (%) |
|---|---|---|---|---|
| **sRNN** | 0.87 | **0.89** | 0.87 | **0.89** |
| **LSTM** | 0.87 | 0.88 | 0.87 | 0.88 |
| **BLSTM** | 0.86 | 0.88 | 0.87 | 0.88 |
| **GRU** | 0.87 | 0.88 | 0.87 | 0.88 |

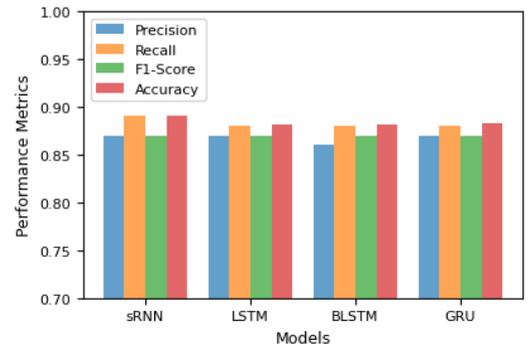

Fig. 3. Classification model performance

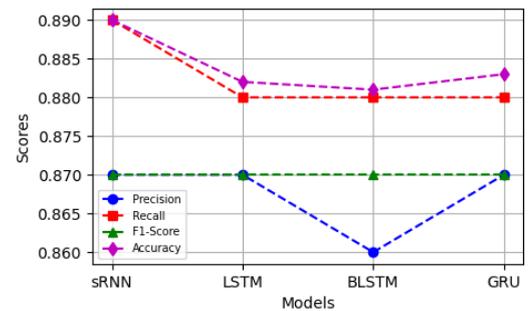

Fig. 4. Performance Comparison of Different Models

As illustrated in Fig. 3 and Fig. 4, the results clearly demonstrate the effectiveness of NLP and Deep Learning models in handling the complexity of aviation safety reports and inferring aircraft damage level information from the textual narratives. Notably, the sRNN model exhibited the highest accuracy and precision, indicating it's suitability for this task.

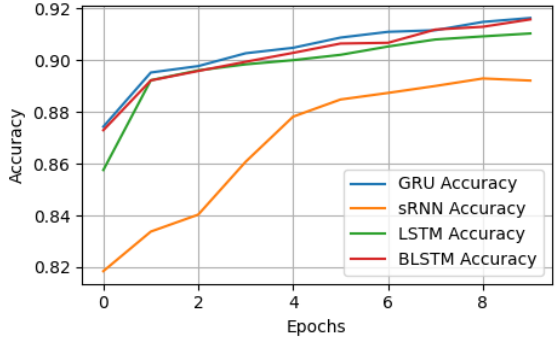

Fig. 5. Validation accuracy performance for all models

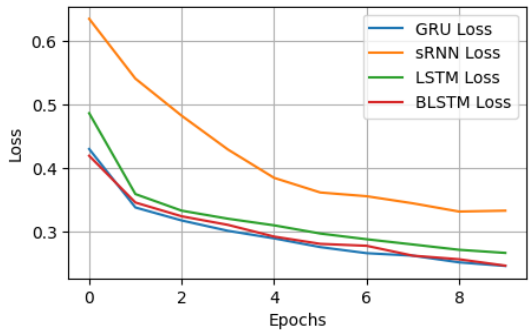

Fig. 6. Validation Loss for all models

Both Fig. 5 and Fig. 6, illustrate the training behavior and performance of the four deep learning models (sRNN, BLTM, LSTM, and GRU) in terms of validation accuracy and validation loss, respectively, over a range of training epochs. These visualizations provide valuable information for assessing and comparing the models' capabilities in solving the classification problem at hand.

```
              Classification Report
              ---------------------
              precision    recall  f1-score   support

       Minor       0.69      0.56      0.61       675
        None       0.91      0.98      0.94      8737
 Substantial       0.93      0.84      0.88       416
   Destroyed       0.64      0.24      0.35       801

    accuracy                           0.89     10629
   macro avg       0.79      0.65      0.69     10629
weighted avg       0.87      0.89      0.87     10629
```

Fig. 7. Classification report for the best model (sRNN)

The extract in Fig. 7 shows the classification report in terms of accuracy, precision, recall, and F1 score for the sRNN model. The extract also gives an account of the test instance distribution among distinct damage level entries as evidenced in the support column. On the other hand, Fig. 8 gives a visual account of how the models distribute test instances in the form of a confusion matrix.

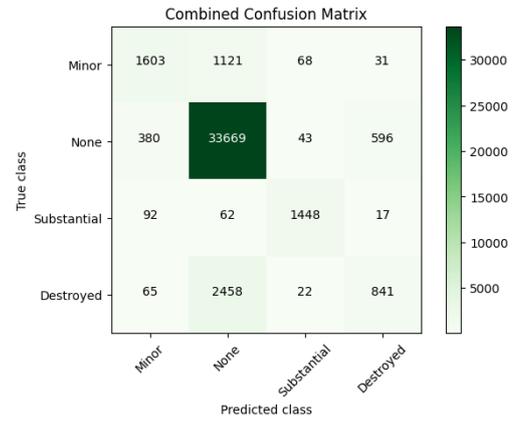

Fig. 8. Confusion Matrix for all of the four models

## V. CONCLUSION

This study examined four deep-learning architectures for the classification of damage levels within aircraft incident narratives sourced from the ATSB dataset. The comparative analysis of sRNN, LSTM, BLSTM, and GRU models for textual damage level classification highlighted the superior performance of the sRNN model in terms of both accuracy and recall. With precision and F1 scores on par with the other models, the notable lead in accuracy and recall underscores the effectiveness of the sRNN architecture in accurately predicting damage levels. Despite the competitive performance of LSTM, BLSTM, and GRU, the dominance of the sRNN model in key evaluation metrics signifies its potential as a reliable solution for textual damage level classification tasks.

The findings presented in Table III, Fig. 3, and Fig. 4 clearly demonstrate the efficiency and accuracy of the deep learning models in handling the complexity of aviation safety reports and inferring aircraft damage level information from the textual narratives. The performance evaluation presented in Figs. 5 and 6 provides valuable insights into the training behavior and capabilities of the models, showcasing their ability for solving the classification problem. Additionally, the classification report for the sRNN model, as depicted in Fig. 7, offers a comprehensive view of the model's precision, recall, and F1 score, shedding light on the test instance distribution across different damage level categories. The corresponding confusion matrix shown in Fig. 8 further visualizes how the models distribute test instances, providing a holistic perspective on their classification performance.

Looking ahead, the exploration of additional research directions holds significant promise. Integration of advanced textual features and semantic analysis techniques could further enhance the models' understanding of nuanced language structures and contextual information within the incident narratives. Also, the incorporation of multimodal data sources, such as image and video data from incident reports, can enrich the classification models' comprehension of the damage severity, enabling a more comprehensive and holistic analysis of the incidents.